\documentclass[10pt,twocolumn,letterpaper]{article}

\usepackage{cvpr2022/cvpr}      

\usepackage{graphicx}
\usepackage{amsmath}
\usepackage{amssymb}
\usepackage{booktabs}
\usepackage{xcolor}
\usepackage{tabularx}
\usepackage{colortbl}
\usepackage{setspace}

\newcolumntype{s}{>{\columncolor[gray]{.85}[.5\tabcolsep]}c}

\usepackage[accsupp]{axessibility}  

\usepackage{gensymb}
\usepackage{multirow}
\usepackage{paralist}
\usepackage[normalem]{ulem}

\usepackage[belowskip=1pt,aboveskip=5pt,font=small]{caption}
\usepackage[belowskip=0pt,aboveskip=3pt,font=small]{subcaption}
\newcommand{\myquote}[1]{\emph{`#1'}}
\newcommand{\myapprox}{{\raise.17ex\hbox{$\scriptstyle\sim$}}}
\newcommand{\xhdr}[1]{\vspace{0pt}\noindent\textbf{#1}\xspace}

\makeatletter
\newcommand\footnoteref[1]{\protected@xdef\@thefnmark{\ref{#1}}\@footnotemark}
\makeatother

\DeclareMathSymbol{@}{\mathord}{letters}{"3B}

\newcommand{\rgbd}{\texttt{RGB-D}\xspace}
\newcommand{\rgb}{\texttt{RGB}\xspace}
\newcommand{\depth}{\texttt{Depth}\xspace}

\newcommand{\pointnav}{\texttt{PointGoalNav}\xspace}

\newcommand{\compassgps}{\texttt{GPS+Compass}\xspace}
\newcommand{\gpscompass}{\compassgps}
\newcommand{\resizeshortestedge}{\texttt{ResizeShortestEdge}\xspace}
\newcommand{\centercrop}{\texttt{CenterCrop}\xspace}

\newcommand{\flipaug}{\texttt{Flip}\xspace}
\newcommand{\swapaug}{\texttt{Swap}\xspace}

\DeclareMathSymbol{@}{\mathord}{letters}{"3B}
\newcommand{\turnleft}{\texttt{turn\_left}\xspace}
\newcommand{\turnright}{\texttt{turn\_right}\xspace}
\newcommand{\rotate}{\texttt{turn\_\{left,right\}}\xspace}
\newcommand{\forward}{\texttt{move\_forward}\xspace}
\newcommand{\callstop}{\texttt{stop}\xspace}

\definecolor{stdevcolor}{HTML}{404040}
\newcommand{\stdev}[1]{\textcolor{stdevcolor}{\footnotesize{$\pm$#1}}}

\renewcommand{\pointnav}{PointNav\xspace}

\renewcommand{\rgbd}{RGB-D\xspace}
\newcommand{\hc}[1]{Habitat Challenge~#1\xspace}
\definecolor{citecolor}{HTML}{2779af}
\definecolor{linkcolor}{HTML}{c0392b}
\usepackage[pagebackref,breaklinks=true,colorlinks,bookmarks=false,citecolor=citecolor,linkcolor=linkcolor]{hyperref}

\usepackage[capitalize]{cleveref}
\crefname{section}{Sec.}{Secs.}
\Crefname{section}{Section}{Sections}
\Crefname{table}{Table}{Tables}
\crefname{table}{Tab.}{Tabs.}

\usepackage{mysymbols}

\def\cvprPaperID{10331} 
\def\confName{CVPR}
\def\confYear{2022}

\begin{document}

\abovedisplayskip 3.0pt plus2pt minus2pt%
\belowdisplayskip \abovedisplayskip

\newcommand{\csection}[1]{
    \vspace{-0.06in}
    \section{#1}
    \vspace{-0.06in}
}

\newcommand{\csubsection}[1]{
    \vspace{-0.06in}
    \subsection{#1}
    \vspace{-0.06in}
}

\newcommand{\csubsubsection}[1]{
    \subsubsection{#1}
}


\title{Is Mapping Necessary for Realistic PointGoal Navigation?}

\author{Ruslan Partsey$^{1}$\thanks{Correspondence to \href{mailto:partsey@ucu.edu.ua}{partsey@ucu.edu.ua}} \ 
Erik Wijmans$^{2,3}$
Naoki Yokoyama$^{2}$
Oles Dobosevych$^{1}$
Dhruv Batra$^{2,3}$
Oleksandr Maksymets$^{3}$\\
\normalsize
$^1$Ukrainian Catholic University
$^2$Georgia Institute of Technology
$^3$Meta AI \\
{\normalsize
\href{https://rpartsey.github.io/pointgoalnav}{rpartsey.github.io/pointgoalnav}
}
}

\maketitle

\begin{abstract}
Can an autonomous agent navigate in a new environment without building an explicit map? 

For the task of PointGoal navigation (\myquote{Go to $\Delta x$, $\Delta y$}) under idealized settings (no \rgbd and actuation noise, perfect GPS+Compass), the answer is a clear `yes' -- map-less neural models composed of task-agnostic components (CNNs and RNNs) trained with large-scale reinforcement learning achieve 100\%~Success on a standard dataset (Gibson~\cite{ramakrishnan2021habitat}). However, for PointNav in a \emph{realistic} setting (\rgbd and actuation noise, no GPS+Compass), this is an open question; one we tackle in this paper. The strongest published result for this task is 71.7\% Success~\cite{zhao2021the}.\footnote{According to \hc{2020} PointNav benchmark held annually. A concurrent as-yet-unpublished result has reported 91\%~Success on 2021's benchmark, but we are unable to comment on the details because an associated report is not available.}

First, we identify the main (perhaps, only) cause of the drop in performance: absence of GPS+Compass. 
An agent with perfect GPS+Compass faced with \rgbd sensing and actuation noise achieves 99.8\%~Success (Gibson-v2 val). This suggests 
that (to paraphrase a meme) robust visual odometry is all we need for realistic PointNav; if we can achieve that, we can ignore the sensing and actuation noise.

With that as our operating hypothesis, we scale dataset and model size, and develop human-annotation-free data-augmentation techniques to train models for visual odometry. We advance the state of art on the Habitat Realistic PointNav Challenge 
from 71\% to 94\% Success (+23, 31\% relative) and 
53\% to 74\% SPL (+21, 40\% relative). 
 While our approach does not saturate or `solve' this dataset, this strong improvement combined with promising zero-shot sim2real transfer (to a LoCoBot)
provides evidence consistent with the hypothesis that explicit mapping may not be necessary for navigation, even in a realistic setting. 


\end{abstract}

\csection{Introduction}
\begin{figure}
    \begin{center}
        \includegraphics[width=1\linewidth]{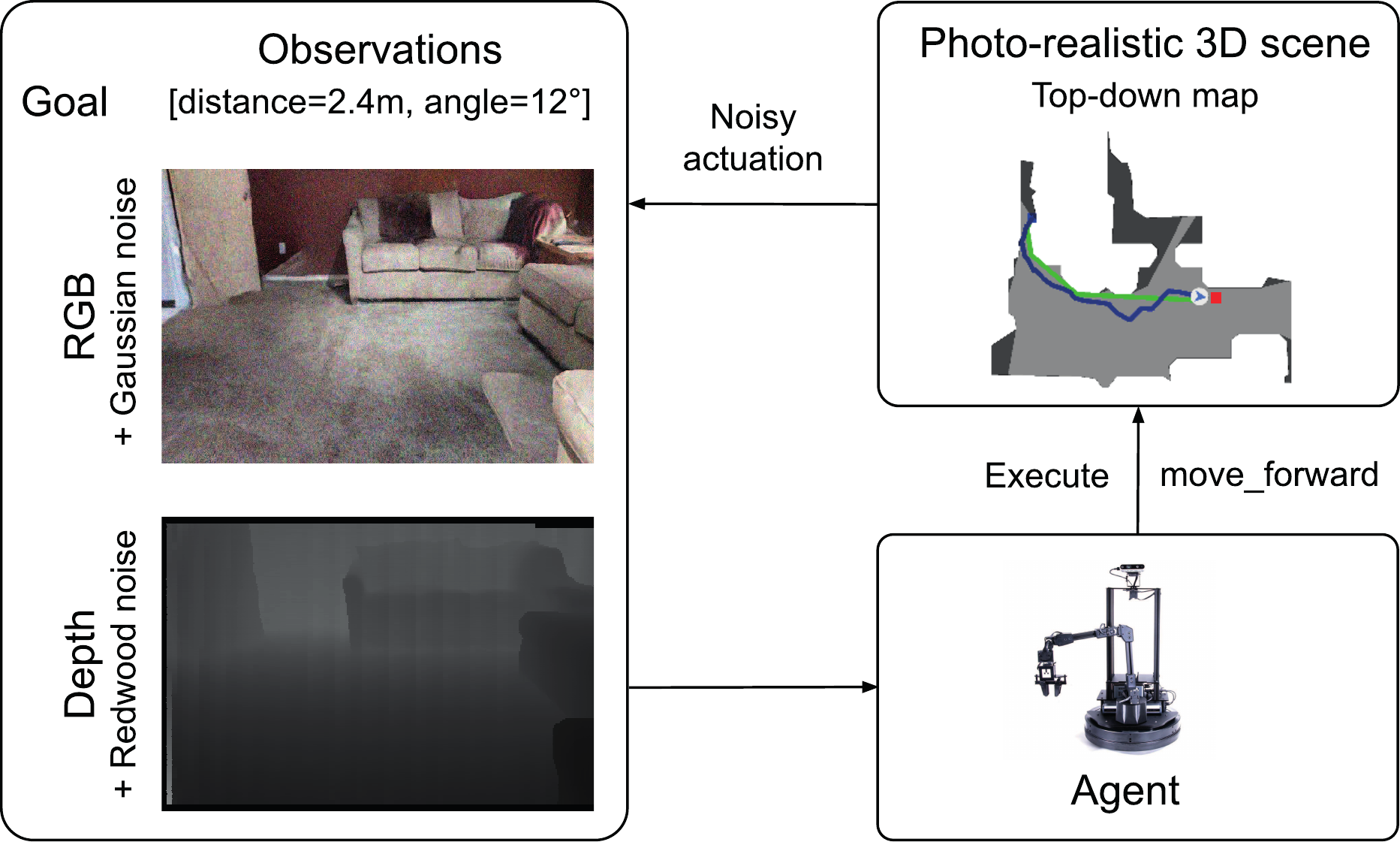}
\end{center}
\vspace{-5pt}
\caption{\xhdr{PointNav.} The agent is tasked with navigating from its starting location (blue square) to a goal location (red square) specified as coordinates relative to its initial place. It must do so from solely an \rgbd camera. The blue line shows the agent’s trajectory, and the green line indicates the oracle path.}
\label{fig:pointnav-task}
\vspace{-5pt}
\end{figure}

The ability to navigate in a novel environment solely from egocentric perception is 
an essential requirement for building intelligent and helpful robots. 
To make progress on this long-term vision, the task of PointGoal navigation (\pointnav)~\cite{anderson2018on}, 
\ie asking a robot to \myquote{go to $(\Delta x, \Delta y)$} relative to its starting location), 
has become a core task. 

We are interested in the question -- can an agent navigate in a new environment without building an explicit map?\footnote{We distinguish an explicit mapping mechanism from implicit spatial understanding that may emerge from building task-specific end-to-end-learned representations. The former is designed, the latter is emergent.}

This question is of deep scientific interest. Decades of research in intelligent animal navigation show that various animals build `cognitive maps'~\cite{tolman48,okeefe1978hippocampus} of their environment. For decades, robotics research has treated explicit mapping and localization~\cite{thrun2005probabilistic,nilsson1984shakey,ayache1988building,smith1990estimating} as integral components in a navigation robot. There are many good reasons to develop mapping technology, but we simply do not know whether mapping is \emph{necessary} for navigation. 
One way to resolve this is to refute the contrapositive -- 
if we demonstrate a map-less approach that can navigate, that will imply that explicit mapping is not necessary for successful navigation.


Under idealized settings -- perfect localization via a noise-free \gpscompass sensor, egocentric sensing via a noise-free \rgbd sensor, and absence of any actuation noise -- map-less navigation models composed of task-agnostic neural components (CNNs and RNNs) trained with large-scale reinforcement learning achieve 100\%~Success~\cite{wijmans2020dd,ramakrishnan2021habitat} on a standard dataset (Gibson~\cite{xia2018gibson}). However, under \emph{realistic} settings -- where the agent must self-localize (\ie no \gpscompass sensor), and must contend with \rgbd sensing noise and actuation noise -- this is an open question; one we tackle in this paper. The strongest published result for this task is 71.7\%~Success~\cite{zhao2021the}. 

To make systematic progress, we first study a simpler version of the realistic setting where the agent is given ground-truth \gpscompass, isolating localization difficulties from the ability to deal with noisy perception and control. While prior work in this setting~\cite{zhao2021the} reported fairly a high success rate (97\%), we significantly sharpen this result and demonstrate near-perfect performance again (99.8\% Success on Gibson \pointnav-v2 val). Our results leave no room for doubt and confirm that the \emph{only} performance bottleneck is the agent's ability to self-localize. 


With this limiting-factor identified, we study the localization, or visual odometry (VO) module. It takes as input two successive observations $O_{t-1}$ and $O_t$ and outputs the relative pose change $(\Delta x, \Delta y, \Delta z, \Delta \theta)$, that is then used to update the location of the goal relative to the robot, which is consumed by the navigation policy. 

We present a series of broadly-applicable modifications that improve agent navigation performance considerably, from 64\%~Success/52\%~SPL to 96\%~Success/77\%~SPL on the Gibson val. 
These modifications are all motivated by the need for robust visual odometry in service of navigation, specifically: 

\begin{compactenum}[1.]
\item \textbf{Action conditioning via action embeddings.} It is important to recognize that our goal is not visual odometry in isolation but in the context of navigation. Specifically, we know what action (\forward $0.25m$, \turnleft $30^{\degree}$ or \turnright $30^{\degree}$)  was executed and should use this information; this observation is not new and has been made in prior work~\cite{zhao2021the}. We find that converting a 1-hot representation of the actions into continuous embeddings and concatenating them to the last two fully-connected layers in the VO network significantly improves performance by +8~Success/+5~SPL.
\item \textbf{Training-time data augmentation.} Data augmentation is one of the most successful methods for regularizing learning techniques~\cite{krizhevsky2017imagenet,zhang2017mixup,yun2019cutmix}. We construct navigation- and odometry-specific augmentations -- \eg when an agent rotates in-place to produce observations $O_{t-1}$ and $O_t$, we can create a new training image that relates $O_t$ and $O_{t-1}$ via the inverse pose and turning action. We also propose a new augmentation called \flipaug (described in \cref{sec:train-time-augs}). Cumulatively, they improve performance by +2~Success/+1~SPL.
\item \textbf{Test-time data augmentation for ensembling.} To improve robustness we perform all augmentations at test-time and aggregate predictions across all combinations. This improves performance by +3~Success/+3~SPL.
\item \textbf{Increased dataset size and model size.} Finally, we study the effects of increasing the odometry dataset size from 500k observation pairs to 1.5M  (+8~Success/+7~SPL) to 5M (+8~Success/+6~SPL), and increasing the model size (+3~Success/+3~SPL). 
\end{compactenum}




\csection{Preliminaries: PointGoal Navigation}
In PointNav (illustrated in \cref{fig:pointnav-task}), an agent is initialized in previously unseen environment and is tasked to reach the goal specified relative to its starting location. 
The action space is discrete and consists of four types of actions: \callstop (to end the episode), \forward by $0.25m$, \turnleft and \turnright by angle $\alpha$\footnote{In PointNav-v1 $\alpha=10^{\degree}$, in PointNav-v2 $\alpha=30^{\degree}$.}.

The agent is evaluated via three primary metrics.
\begin{inparaenum}[1)]
\item Success, $S_i$, where an episode $i$ is considered  successful  if  the  agent  issues the \callstop command within $0.36m$ (2$\times $agent radius) of the goal.
\item Success weight by (inverse normalized) Path Length (SPL)~\cite{anderson2018on}, where success is weighted by the efficiency of the agent's path. Formally, for episode $i$, let $S_i$ be a binary indicator of success, $p_i$ be the length of the agent's path, and $l_i$ be the length of the shortest path (geodesic distance), then for $N$ episodes
\begin{equation} 
\label{eq:spl} 
\text{SPL} = \frac{1}{N}\sum_{i=1}^{N}S_i \cdot \frac{l_i}{\max(p_i, l_i)}.
\end{equation}
\item  SoftSPL~\cite{datta2020integrating}, where binary success is replaced by progress towards goal. Formally, for episode $i$, let $d_{0_i}$ be the initial distance to goal and $d_{T_i}$ be the distance to goal at the end of the episode (on both successes and failures), then 
\begin{equation} 
\label{eq:softspl} 
\text{SoftSPL} = \frac{1}{N}\sum_{i=1}^{N}\left(1 - \frac{{d_{T_i}}}{{d_{{0}_i}}}\right)\left(\frac{l_i}{\max(p_i, l_i)}\right). 
\end{equation}
\end{inparaenum}

\xhdr{Embodiment.} 
With an eye on sim2real generalization, the agent's specification matches the LoCoBot's \footnote{LoCoBot is a low-cost mobile manipulator suitable for both navigation and manipulation (\href{http://www.LoCoBot.org/}{http://www.LoCoBot.org}).} specification. The agent is equipped with an RGB-D camera mounted at a height of $0.88m$ and tilted $-20^{\degree}$ (angled downwards towards the floor; pitch or camera azimuth angle). Camera’s resolution is 360 $\times$ 640  pixels with $70^{\degree}$ horizontal field of view. Base radius is 0.18m.

\csubsection{PointNav-v1: Idealized (Noise-less) Setting}

In idealized setting (named `v1'), the agent was equipped with noise-free \rgbd camera, given access to ground-truth localization (via an \gpscompass sensor), and movement was deterministic/noise-free (meaning \turnright~$10^{\degree}$ always turned the agent \emph{exactly}~$10^{\degree}$). The agent could also `slide' along walls -- a commonplace behavior in video games that improves human control but was later found to degrade sim-to-real performance~\cite{kadian2019are}. 


State-of-the-art approaches for this task rely on large-scale reinforcement learning and have begun to saturate the available datasets: \eg Wijmans~\etal~\cite{wijmans2020dd} reported 99\%~Success, 94\% SPL on Gibson test, Ramakrishnan~\etal~\cite{ramakrishnan2021habitat} sharpened this result to 100\% Success, 94\% SPL on Gibson test, 94\% Success, 83\% SPL on MP3D test, and 99\%~Success, 92\% SPL on HM3D test. Overall, PointNav-v1 is largely considered satured or `solved'. 

\csubsection{PointNav-v2: Realistic (Noisy) Setting}
\label{sec:prelim-v2}
Noiseless sensing and actuation simply do not yet exist. Different lightning conditions, surface properties (such as friction), and other sources of error cause actuation and sensing noise that introduce significant drift over a long trajectory. Moreover, high-precision localization in indoor environments can not be assumed in realistic settings. 

The so-called `realistic' (or v2) setting of PointNav addresses these shortcomings of the v1 by introducing actuation noise (modeled by benchmarking the LoCoBot robot~\cite{murali2019pyrobot}), removing \gpscompass, and adding noise to the \rgbd camera. To simulate real-world camera RGB and Depth, noise models from~\cite{choi2015robust} were used (Gaussian noise model for RGB and Redwood noise model for Depth).

Initial attempts to directly apply PointNav-v1 techniques to PointNav-v2 were largely unsuccessful ($\approx$5\%~Success~\cite{datta2020integrating}). More recent methods~\cite{datta2020integrating,zhao2021the} train a navigation policy with access to ground-truth localization and then replace ground-truth localization with estimated localization by integrating the egomotion estimates from a visual odometry module. The strongest published result for this task is 71.7\%~Success and 53\%~SPL~\cite{zhao2021the}, indicating that navigation with noisy actuation and sensing  continues to be an open frontier for research.




\csection{Navigation Policy}
\begin{figure}[t!]
    \begin{center}
        \includegraphics[width=1\linewidth]{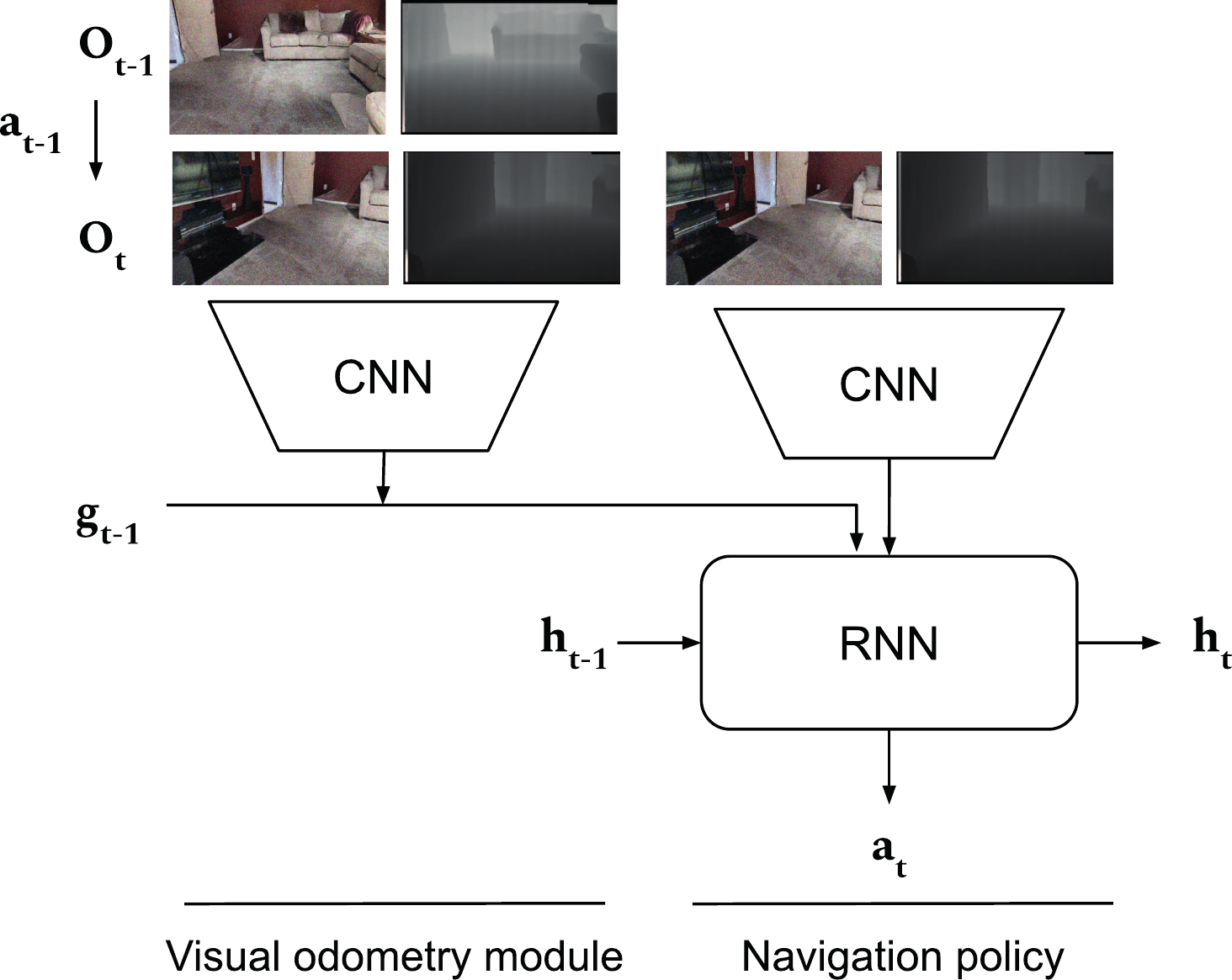}
\end{center}
\vspace{-5pt}
\caption{\xhdr{Agent architecture for Realistic PointGoal navigation} consisting of an RNN-based RL navigation policy and CNN-based visual odometry (VO) module. Inputs are  $g_{t-1}$ - goal coordinates wrt.~previous pose, $a_{t-1}$ - previous action, $O_{t-1}$ - observations at previous timestep, and $O_t$ - current observations. First, VO predicts the change between $t-1$ and $t$ and then update the goal to be wrt.~current pose. The updated goal location is given to the navigation policy along with $O_t$ to predict the next action $a_t$. The initial goal location estimate is equal to ground truth goal location (as per the task specification).}
\label{fig:policy_vo}
\vspace{-5pt}
\end{figure}

Our pipeline consists of two components: a navigation policy~(nav-policy) that given observations $O_t$ at time step~$t$ decides which action to take to reach the goal and a visual odometry (VO) module that given two consecutive observations $(O_{t-1}$, $O_{t})$ estimates relative pose change (egomotion) that is further used to update the goal coordinates after each step (see \cref{fig:policy_vo}). This decoupling of roles is a natural choice. It builds upon the results in the idealized setting that has been used in prior work and is used in the previous state of the art~\cite{datta2020integrating,zhao2021the}. In this section we describe our navigation policy and show that it is capable of near-perfect navigation with noisy actuations and noisy \rgbd sensing when given ground-truth localization, demonstrating that visual odometry is the bottleneck. We describe the details of our visual odometry module in the next section (\cref{section:vo}). We use the Habitat platform~\cite{savva2019habitat, szot2021habitat} to simulate navigation experiments.

\csubsection{Architecture}
We train the navigation assuming perfect odometry (using ground-truth localization provided by \gpscompass sensor) and then use VO module estimates as a drop-in replacement of ground-truth localization sensor without fine-tuning (idea introduced by Datta \etal~\cite{datta2020integrating}; also used by Zhao \etal~\cite{zhao2021the}). This also allows us to evaluate the policy's performance in isolation of the visual odometry module.

We use the same  navigation policy as Wijmans~\etal~\cite{wijmans2020dd}, consisting of a two-layer Long Short-Term Memory (LSTM)~\cite{hochreiter1997long} and a half-width ResNet50~\cite{he2016deep} encoder. At each timestep, the policy is given the output from the noisy \depth sensor (following common practice for the navigation policy, we discard \rgb) and idealistic \gpscompass sensor (ground-truth localization, that is replaced by the visual odometry estimates during evaluation). Before passing through the feature encoder, visual observations are transformed using \resizeshortestedge and \centercrop observation transforms; the former resizes the shortest edge of the input to 256 pixels while maintaining aspect ratio, the latter center crops the input to 256~$\times$~256 pixels. 

\csubsection{Training Details}

\begin{figure*}[ht!]
\begin{center}
\includegraphics[width=1\linewidth]{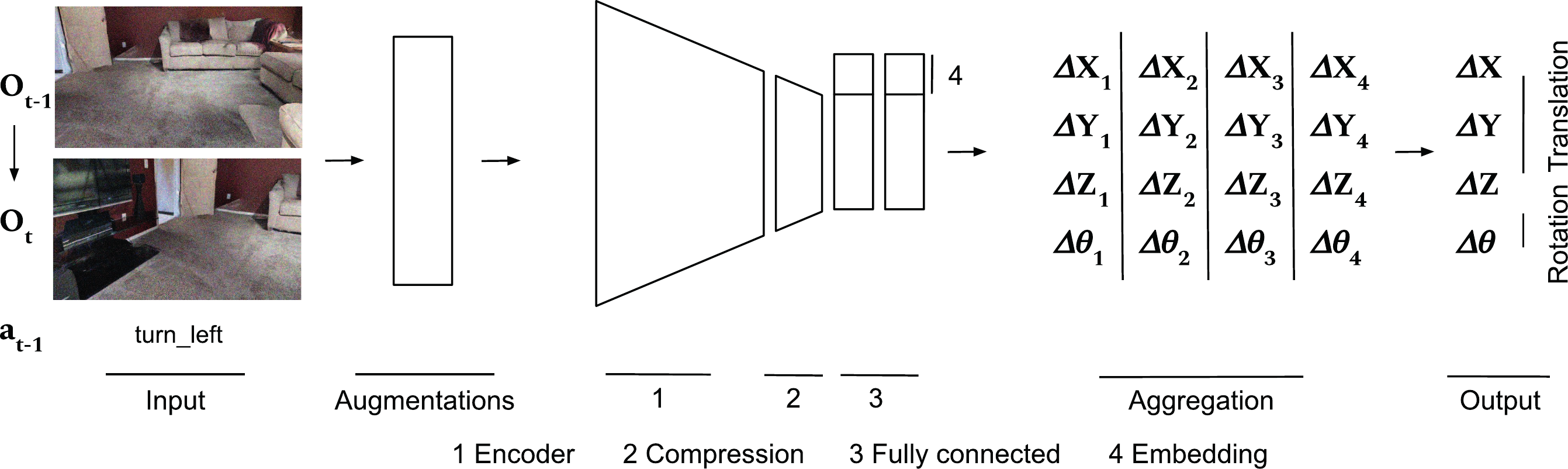}
\end{center}
\caption{\xhdr{Visual odometry module.} At inference the input pair of observations $(O_{t-1}, O_t)$ is transformed by applying \swapaug and \flipaug augmentations. In total the visual odometry model receives two observation pairs for \forward (original and flipped) and four observation pairs for \rotate actions (original, flipped, swapped(original), swapped(flipped)). In the aggregation stage outputs are transformed back to original coordinate frame by applying the inverse transformation for each augmentation and then averaging to produce the final egomotion estimate (details in \cref{sec:test-time-augs}).}
    \label{fig:vo_module}
\vspace{-5pt}
\end{figure*}

We use the train split of the full Gibson data~\cite{xia2018gibson} (scans with ratings of 0+ from \cite{savva2019habitat}).
We leverage Decentralized Distributed Proximal Policy Optimization (DD-PPO)~\cite{wijmans2020dd} and Wijmans~\etal's reward structure to train the policy. 

For an episode $i$, the agent receives a `terminal' reward of $r_T = 2.5 \cdot \text{Success}_i$ ($r_T = 2.5 \cdot \text{SPL}_i$ in later experiments) that encourages it to stop at the correct location (and take an efficient path), and a shaped reward $r_t(a_t, s_t) = -\Delta_{\text{geo\_dist}} - 0.01$, that encourages it to take steps towards the goal (while being efficient), where $\Delta_{\text{geo\_dist}}$ is the change in geodesic distance to the goal by performing action $a_t$ in state $s_t$. Note that reward is not available at test-time. We train with 64 GPUs (workers). 
We trained for 2.5 billion steps on Gibson~4+, then for another 2.5 billion steps on Gibson~0+, and finally for another 2.5 billion steps on Gibson~0+ with the terminal reward weighted by SPL. We started each stage with the best (by val performance) agent from the previous stage.

\begin{figure*}
\begin{center}
\includegraphics[width=1\linewidth]{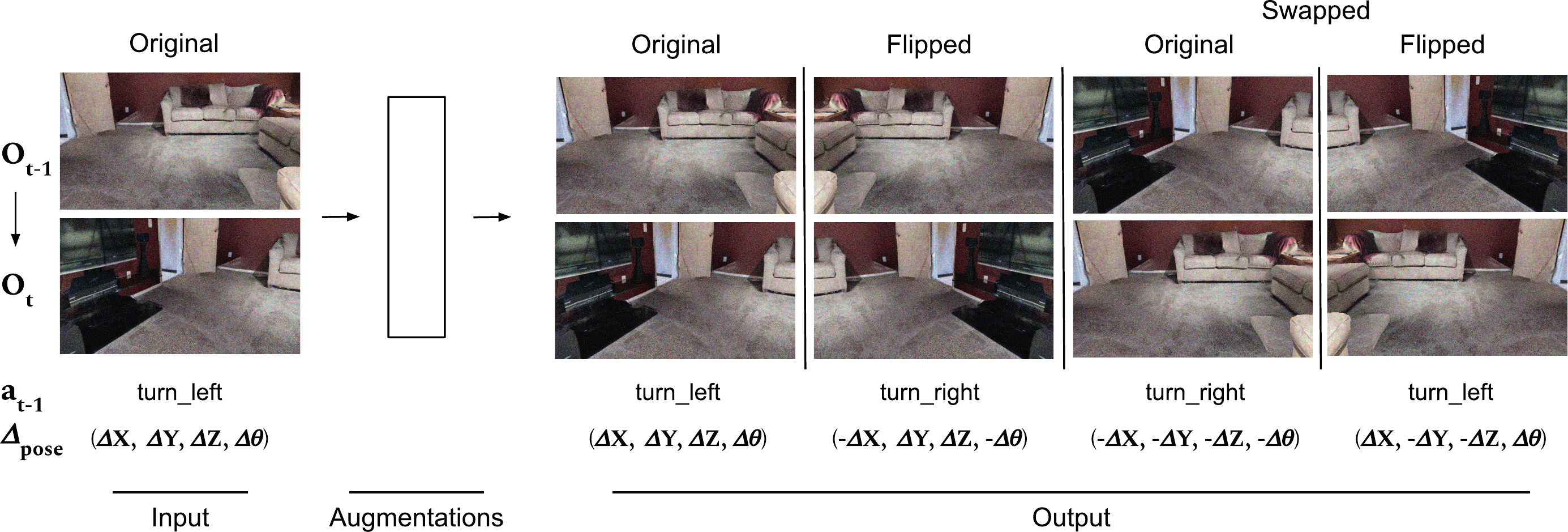}
\end{center}
\caption{\xhdr{Augmentations.} An input pair of observations $(O_{t-1}, O_t)$ are transformed by applying \swapaug (returns $(O_t, O_{t-1}$) and \flipaug (flips observations about their vertical axis) augmentations, producing two transformed pairs as output for \forward action (original and flipped) and four transformed pairs as output for \rotate actions (original, flipped, swapped(original), swapped(flipped)) (details in \cref{sec:train-time-augs}).}
\label{fig:vo_augs}
\end{figure*}

\csubsection{Ground-Truth Localization Performance}
\label{sec:gt-loc-perf}

Our agent achieves 99.8\%~Success and 80\% SPL on Gibson val PointNav-v2 dataset. 
This result shows that   
near-perfect Success is achievable without building an explicit map 
if we can assume perfect odometry,  
even with noisy observations and actuations. 

Note that the SPL (80\%) is \emph{not} near-perfect by any 
reasonable definition of `near'. However, let us ask -- \emph{is 100\% SPL actually achievable?} 
Recall that actuations are noisy. Even an oracle agent with full knowledge of the environment may not be able to follow the shortest path and achieve 100\% SPL because the noisy actuation may take it off the shortest path. 
This problem is exacerbated because `sliding' is disabled in v2, meaning that if an agent is traveling close to an obstacle (as shortest path typically do), noisy actuation may bring it into contact with the obstacle, requiring backtracking or dislodging and adding to its path length.

To determine an upper-bound on SPL in \pointnav-v2, we implement a heuristic planner that uses the ground-truth map to choose motion primitives (\rotate$\times$\texttt{N}, then \forward). The planner selects the primitive that best reduces distance to goal using the ground-truth geodesic distance (computed using the ground-truth map), executes the first action in the selected primitive, and then re-runs the selection process until the goal is reached. On Gibson val, the oracle achieves 84\% SPL. This confirms that we should not expect 100\% SPL in \pointnav-v2.

While the gap between this oracle (84\% SPL) and our agent's performance (80\% SPL) continues to be non-trivial, it is significantly smaller 
than the best known-result without perfect odometry (63\% SPL) \cite{zhao2021the}. 
Thus, it is clear that 
that visual odometry is the limiting factor, 
and we turn our focus towards this component for the rest of the paper.

\csection{Visual Odometry}
\label{section:vo}

The visual odometry model takes a pair of 180$\times$360 RGB-D frames stacked channel-wise as input and predicts the relative pose change between the two camera positions, $\Delta_\text{pose} = (\Delta x, \Delta y, \Delta z, \Delta \theta)$ , where $\Delta x, \Delta y, \Delta z$ refer to the 3D translation of the camera center and $\Delta \theta$ refers to the rotation about the gravity vector (`yaw' or robot heading). 

The visual odometry model is represented as ResNet~\cite{he2016deep} encoder followed by a compression block and two fully connected (FC) layers. 
We replace BatchNorm~\cite{ioffe2015batch} with  GroupNorm~\cite{wu2018group}  (we found it to work better) and use half of the width. 
The compression block consists of 3$\times$3 Conv2d+GroupNorm+ReLU. We apply DropOut~\cite{srivastava2014dropout} with 0.2 probability between fully connected layers. 
Full VO pipeline is illustrated in \cref{fig:vo_module}. 

\csubsection{Action Embedding}
Actuation noise and collisions affect agent translation and rotation for each action type differently (the agent may rotate while moving forward and move while rotating in place~\cite{murali2019pyrobot}). This motivated us to study incorporating knowledge of the action taken between two consecutive observations as an additional input.
We represent the action taken as an embedding -- a fixed/frozen action-specific vector of length 16 that is concatenated to the flattened output from the feature encoder. We \emph{do not} apply a DropOut to action embedding as we find this harms performance. To further increase the importance of the action, we concatenated the embedding to the input of all fully connected layers.

\csubsection{Train-Time Augmentations}
\label{sec:train-time-augs}
Given a pair of observations, $(O_{t-1}, O_t)$, the action taken between them, $a_{t-1}$, and their relative change in pose $\Delta_\text{pose}$, we use the following augmentations.

\xhdr{\swapaug.} For every training tuple $(O_{t-1}, O_t, a_{t-1}, \Delta_\text{pose})$ where $a_{t-1}$ is a rotation action we create an extra training example $(O_{t}, O_{t-1}, a^\swapaug_{t-1}, \Delta^\swapaug_\text{pose})$ where $a^\swapaug_{t-1}$ is the effective action taken (\turnleft$\rightarrow$~\turnright and \turnright$\rightarrow$~\turnleft) and $\Delta^\swapaug_\text{pose}$ is the change in pose after swapping (\ie, negation of all components). As in Zhao \etal~\cite{zhao2021the}, this augmentation leverages the order invariance of \rotate actions.

\xhdr{\flipaug.} In architecture and indoor design, it is common to prepare mirror-image floor plans (\eg, kitchen on the left, living on the right and vice versa) to increase the number of options. As shown in \cref{fig:vo_augs}, we simulate the robot navigating in a mirror-image house by flipping its camera image about the vertical axis. Specifically, for every training tuple $(O_{t-1}, O_t, a_{t-1}, \Delta_\text{pose})$ we generate an additional training example $(O^\flipaug_{t-1}, O^\flipaug_{t}, a^\flipaug_{t-1}, \Delta^\flipaug_\text{pose})$, where $O^\flipaug_{t-1}$, $O^\flipaug_{t}$ are the \rgbd observations flipped along their vertical axis, $a^\flipaug_{t-1}$ is the effective action after the flip (\turnleft$\rightarrow$~\turnright, \turnright$\rightarrow$~\turnleft, and \forward remains the same), and $\Delta^\flipaug_\text{pose} = (-\Delta x, \Delta y, \Delta z, -\Delta \theta)$.

We also apply the composition of \flipaug and \swapaug (similar to \texttt{torchvision.transforms.Compose}~\cite{NEURIPS2019_9015}). Note that
the two operations are commutative, \ie, 
\flipaug then \swapaug is the same as \swapaug then \flipaug. All four combinations of Flip and Swap are shown in \cref{fig:vo_augs}. 


\begin{table*}
\begin{center}
\resizebox{\textwidth}{!}{
\begin{tabular}{@{}ccccccccccsccc@{}}
\toprule
\multicolumn{1}{c}{} & 
\multicolumn{1}{c}{Dataset} &
\multicolumn{2}{c}{VO} &
\multicolumn{2}{c}{Embedding} &
\multicolumn{2}{c}{Train time} &
\multicolumn{2}{c}{Test time} &
\multicolumn{4}{c}{Navigation metrics ($\times10^2$)} 
\\
\cmidrule(lr){3-4} \cmidrule(lr){5-6} \cmidrule(lr){7-8} \cmidrule(lr){9-10} \cmidrule(l){11-14}
\multicolumn{1}{l}{} &
\multicolumn{1}{c}{size(M)} &                               
\multicolumn{1}{c}{Encoder} & 
\multicolumn{1}{c}{Size(M)} & 
\multicolumn{1}{c}{1FC} &
\multicolumn{1}{c}{2FC} &
\multicolumn{1}{c}{Flip} &
\multicolumn{1}{c}{Swap} &
\multicolumn{1}{c}{Flip} &
\multicolumn{1}{c}{Swap} &
\multicolumn{1}{c}{Success} &
\multicolumn{1}{c}{SPL} &
\multicolumn{1}{c}{SoftSPL} &
\multicolumn{1}{c}{$\text{d}_\text{G}$} 
\\ 
\midrule
1 & 0.5 & ResNet18 &  4.2 & & & & & &            
& 64.0\stdev{0.9} & 51.9\stdev{1.0} & 72.6\stdev{0.3} & 57.7\stdev{2.5} 
\\ 
2 & 0.5 & ResNet18 &  4.2 & \checkmark & & & & &            
& 70.6\stdev{1.9} & 56.5\stdev{1.7} & 73.3\stdev{0.3} & 43.7\stdev{1.7} 
\\ 
3 & 0.5 & ResNet18 &  4.2 & \checkmark & \checkmark & & & &            
& 72.1\stdev{1.0} & 57.5\stdev{0.4} & 73.4\stdev{0.5} & 42.7\stdev{2.6} 
\\ 
\midrule
4 & 0.5 & ResNet18 &  4.2 & \checkmark & \checkmark & & \checkmark & & 
& 69.5\stdev{1.3} & 55.6\stdev{1.1} & 72.5\stdev{0.2} & 50.6\stdev{1.7} 
\\
5 & 0.5 & ResNet18 &  4.2 & \checkmark & \checkmark & & \checkmark & & \checkmark
& 71.0\stdev{1.3} & 57.0\stdev{1.2} & 72.5\stdev{0.5} & 53.6\stdev{2.4} 
\\ 
\midrule
6 & 0.5 & ResNet18 &  4.2 & \checkmark & \checkmark & \checkmark & & & 
& 73.5\stdev{1.8} & 58.7\stdev{1.3} & 74.0\stdev{0.1} & 39.7\stdev{1.9} 
\\
7 & 0.5 & ResNet18 &  4.2 & \checkmark & \checkmark & \checkmark & & \checkmark & 
& 75.7\stdev{0.8} & 60.7\stdev{0.8} & 74.1\stdev{0.2} & 37.1\stdev{2.3} 
\\ 
\midrule
8 & 0.5 & ResNet18 &  4.2 & \checkmark & \checkmark & \checkmark & \checkmark & & 
& 73.7\stdev{0.3} & 58.8\stdev{0.6} & 72.8\stdev{0.2} & 45.2\stdev{2.9} 
\\
9 & 0.5 & ResNet18 &  4.2 & \checkmark & \checkmark & \checkmark & \checkmark & & \checkmark 
& 75.5\stdev{1.2} & 60.3\stdev{0.9} & 73.5\stdev{0.2} & 40.3\stdev{1.9} 
\\ 
10 & 0.5 & ResNet18 &  4.2 & \checkmark & \checkmark & \checkmark & \checkmark & \checkmark & 
& 76.2\stdev{0.9} & 60.6\stdev{0.8} & 73.3\stdev{0.3} & 39.2\stdev{2.0} 
\\
11 & 0.5 & ResNet18 &  4.2 & \checkmark & \checkmark & \checkmark & \checkmark & \checkmark & \checkmark 
& 77.0\stdev{0.8} & 61.5\stdev{0.5} & 73.9\stdev{0.3} & 37.2\stdev{1.0} 
\\ 
\midrule 
12 & 1.5 & ResNet18  & 4.2 & \checkmark & \checkmark &            &            &            &            
& 77.0\stdev{1.3} & 62.0\stdev{1.0} & 74.3\stdev{0.3} & 38.0\stdev{0.9} 
\\ 
13 & 1.5 & ResNet18  & 4.2 & \checkmark & \checkmark & \checkmark & \checkmark &            &  
& 80.0\stdev{0.9} & 64.1\stdev{0.9} & 73.8\stdev{0.4} & 37.8\stdev{0.7} 
\\
14 & 1.5 & ResNet18 & 4.2 &   \checkmark & \checkmark & \checkmark & \checkmark & \checkmark & \checkmark 
& 85.2\stdev{0.5} & 68.4\stdev{0.2} & 74.9\stdev{0.4} & 31.5\stdev{1.5} 
\\ 
\midrule 
15 & 1.5 & ResNet50 &  7.6 & \checkmark & \checkmark & \checkmark & \checkmark & \checkmark & \checkmark
& 88.0\stdev{0.6} & 70.6\stdev{0.2} & 75.5\stdev{0.2} & 26.8\stdev{1.7} 
\\
\midrule 
16 & 5 & ResNet50 &  7.6 & \checkmark & \checkmark & \checkmark & \checkmark & \checkmark & \checkmark
& 96.0\stdev{0.5} & 76.6\stdev{0.4} & 76.4\stdev{0.3} & 20.1\stdev{0.8}  
\\
\midrule 
17 & \multicolumn{9}{c}{Ground-Truth Odometry}
& 99.8\stdev{0.1} & 79.8\stdev{0.2} & 77.0\stdev{0.2} & 16.2\stdev{1.1} 
\\
\bottomrule
\end{tabular}
}
\end{center}
\vspace{-5pt}
\caption{\xhdr{Evaluation on the Gibson v2 4+ validation split.}
Results are reported as an average of four evaluations with different seeds. Navigation metrics Success, SPL, SoftSPL, $\text{d}_\text{G}$ (distance to goal) are subject to $\times10^2$ multiplication. Tick in a column indicate whether a particular option is turned on. For instance, in row 6 ResNet18 is the VO encoder, action embedding is concatenated to 1-st fully connected and 2-nd fully connected layer, flip augmentation is turned on during training, and navigation metrics are reported with no augmentations during evaluation.}
\label{table:nav_metrics}
\end{table*}

\csubsection{Test-Time Augmentations}
\label{sec:test-time-augs}

We adapt the common practice of test-time augmentation in image classification to visual odometry.
Specifically, we apply \flipaug and \swapaug augmentations during the test time (\ie during navigation) then aggregate pose-predictions. The aggregation consists of two steps: first we transform egomotion estimates for transformed input pairs back to original coordinate system, $\flipaug^{-1}(\Delta_\text{pose}) = (-\Delta x, \Delta y, \Delta z, -\Delta \theta)$,  $\swapaug^{-1}(\Delta_\text{pose}) = -\Delta_\text{pose}$, and then take the average (illustrated in \cref{fig:vo_module}).


\csubsection{Training Details}
\label{sec:vo_training_details}
We train the visual odometry model decoupled from the navigation policy -- on a static dataset $\calD = \{ (O_{t-1}, O_t, a_{t-1}, \Delta_\text{pose}) \}$. This dataset is created by using the oracle to unroll trajectories from which the pairs of RGB-D frames with meta-information about actions taken and egomotions are uniformly sampled (similar dataset collection protocol were used by~\cite{datta2020integrating,zhao2021the}). We use Gibson 4+ scenes (and Gibson-v2 PointGoal navigation episodes) to generate the VO dataset. We collect the training dataset by uniformly sampling 20\% of pairs of observations from train scenes (500k to 5M total training tuples)  and the validation dataset by sampling 75\% of pairs of observations from validation scenes (34k total). 

The model is trained with batch size 32, Adam optimizer with learning rate $10^{-4}$ and mean squared error (MSE) loss for both translation and rotation.

\csubsection{Dataset and Model Size}
We vary the dataset that the VO module is trained on from 500k to 5M observation tuples and experiment with ResNet18 and ResNet50 encoders.

As the training time increases linearly to the dataset size, we also implemented the distributed VO training pipeline that allows for multi-node multi-GPU scaling and significantly reduces experiment time. Training on 8 nodes (with 8 GPUs each) runs 6.4 $\times$ faster that training on 1 node.






\csection{Experiments}

We report experiments results in \cref{table:nav_metrics}.
Experiments in rows 1-15 were run for 50 epochs and 90 epochs for row 16 
(our best performing agent). 
We perform early-stopping via validation loss. To study the impact of different visual odometry modules we fixed the navigation policy (used the same network weights) across all experiments. 

\csubsection{Ablations}
In this section we study the importance of proposed additions over a baseline VO model: incorporating meta-information available to the agent by adding action embeddings, \flipaug and \swapaug, and larger datasets.
We start from a baseline ResNet18 model (\cref{table:nav_metrics}, row 1).

\xhdr{Action embedding.}
We analyze two possible ways of incorporating meta-information: concatenating the embedding to the \emph{first} FC layer that goes after encoder (\cref{table:nav_metrics}, row~2) and concatenating the embedding to \emph{all} FC layers (row~3). Concatenating action embedding to first FC layer improves performance by +7~Success/+5~SPL compared to a baseline (row~2 vs row~1). Concatenating action embedding to all FC layers improves performance further by +1~Success/+1~SPL (row~3 vs row~2). 
We believe this allows the FC layers to receive more context to learn more accurate egomotion for each action type using shared encoder.





\xhdr{Train-time augmentations.}
Enhancing the VO dataset by applying \flipaug improves performance by +2~Success/+1~SPL (row 6 vs row 3). Interestingly, we find that \swapaug hurts performance by -2~Success/-2~SPL (row 4 vs row 3) while \flipaug+\swapaug achieves performance equivalent to \flipaug (row 8 vs row 6).



\xhdr{Test-time augmentations.}
The biggest performance boost from augmentations comes when they are also applied at test-time (navigation). 
Turning \flipaug on at test-time improves performance by +2~Success/+2~SPL compared to a model with \flipaug on only at train-time (row 10 vs row 6). With \swapaug on at train- and test-time, performance is still worse than achieved by model without \swapaug, -1~Success/-1~SPL (row 5 vs row 3). However, when both \swapaug and \flipaug are on at train- and test-time performance improves further by +1~Success/+1~SPL compared to model with \flipaug  (row 11 vs row 10). That is a total improvement of +5~Success/+4~SPL compared to a model trained and evaluated without augmentations (row 11 vs row 3).

\xhdr{Larger dataset.}
To study the impact of large scale training we increased the training dataset size 3$\times$ (from 500k to 1.5M training pairs) following the same dataset collection protocol, described in \cref{sec:vo_training_details}). Without augmentations, increasing dataset size 3$\times$ improves performance by +5~Success/+4~SPL (row 12 vs row 3) and by +8~Success/+4~SPL(row 14 vs row 11) with augmentations. 

We also examine the impact of augmentations with this larger dataset. Surprisingly, we find that they are \emph{more} influential with a larger training dataset. At train-time, \swapaug+ \flipaug improve performance by +2~Success/+1~SPL with a small dataset (row 8 vs row 3) while they improve performance by +3~Success/+2~SPL (row 13 vs row 12) with a large dataset. A test-time, \swapaug+ \flipaug improve performance by +3~Success/+3~SPL (row 11 vs row 8) with a small dataset  while they improve performance by +5~Success/+4~SPL (row 14 vs row 13) with a large dataset.

\xhdr{Deeper encoder.}
We find that training with more sophisticated encoder architecture (ResNet50 instead of ResNet18) improves navigation performance further by +3~Success/+3~SPL (row 15 vs row 14). Given the additional representational capacity of ResNet50, we further increase the training dataset size to 5M pairs. This improves performance by +8~Success/+6~SPL (row 16 vs row 15).

\xhdr{Dataset transfer.} We examine how the two components of our agent transfer from their training dataset, Gibson, to the Matterport3D dataset~\cite{chang2017mp3d}. We find that while the performance of the agent with ground-truth localization is reduced by a relatively small amount, -6~Succes/-6~SPL (\cref{table:cross-dataset-eval}, row 5 vs row 2), the performance of the agent with visual odometry is reduced by considerably more, -19~Success/-18~SPL (row 6 vs row 3). This observation is analogous to known results in \pointnav-v1, where Depth-only agents (analogous to our agents with known odometry) transfer well from Gibson to Matterport3D, but agents with \rgbd (analogous to our agent with visual odometry) transfer poorly~\cite{savva2019habitat,ramakrishnan2021habitat}. This leaves the question -- is there a universal (cross-dataset) VO module? 
We anticipate creating one will require training on multiple large-scale datasets.

\begin{table}
\begin{center}
\resizebox{\linewidth}{!}{
\begin{tabular}{@{}cccsccc@{}}
\toprule
\multicolumn{1}{c}{} &
\multicolumn{1}{c}{\multirow{2}{*}{Dataset}} & 
\multicolumn{1}{c}{\multirow{2}{*}{Policy}} & 
\multicolumn{4}{c}{Navigation metrics ($\times10^2$)} 
\\
\cmidrule(l){4-7}
\multicolumn{3}{c}{} & 
\multicolumn{1}{c}{Success} &
\multicolumn{1}{c}{SPL} &
\multicolumn{1}{c}{SoftSPL} &
\multicolumn{1}{c}{$\text{d}_\text{G}$} 
\\
\midrule
1 & Gibson & Oracle     & 98.6\stdev{0.1} & 84.5\stdev{0.1} & 80.5\stdev{0.2} & 30.6\stdev{1.5} \\ 
2 & Gibson & Learned+GT & 99.8\stdev{0.1} & 79.8\stdev{0.2} & 77.0\stdev{0.2} & 16.2\stdev{1.1} \\
3 & Gibson & Learned+VO & 96.0\stdev{0.5} & 76.6\stdev{0.4} & 76.4\stdev{0.3} & 20.1\stdev{0.8} 
\\ 
\midrule
4 & MP3D & Oracle       & 98.7\stdev{0.4} & 85.4\stdev{0.4} & 83.2\stdev{0.2} & 34.3\stdev{0.9} \\ 				
5 & MP3D & Learned+GT   & 94.4\stdev{0.6} & 71.8\stdev{0.7} & 70.7\stdev{0.6} & 123.5\stdev{8.7} \\ 				
6 & MP3D & Learned+VO   & 79.4\stdev{1.7} & 60.9\stdev{1.3} & 69.1\stdev{0.3} & 142.9\stdev{16.6} %
\\
\bottomrule
\end{tabular}
}
\end{center}
\vspace{-5pt}
\caption{\xhdr{Dataset transfer.} We evaluate how well the components of our agent transfer from its training dataset (Gibson v2) to the validation dataset of Matterport3D v2. We find that while the policy transfers well, visual odometry performance suffers.} \label{table:cross-dataset-eval}
\vspace{-5pt}
\end{table}

\csubsection{Habitat Challenge 2021 PointNav Track}
\begin{table}
\begin{center}
\resizebox{\linewidth}{!}{
\begin{tabular}{@{}clscccc@{}}
\toprule
\multicolumn{1}{c}{\multirow{2}{*}{Rank}} & 
\multicolumn{1}{c}{\multirow{2}{*}{Participant team}} &
\multicolumn{4}{c}{Navigation metrics ($\times10^2$)} 
\\
\cmidrule(l){3-6}
\multicolumn{2}{c}{} &
\multicolumn{1}{c}{Success} &
\multicolumn{1}{c}{SPL} &
\multicolumn{1}{c}{SoftSPL} &
\multicolumn{1}{c}{$\text{d}_\text{G}$} 
\\
\midrule
1 & VO for Realistic PointGoal (Ours)                              & \textbf{94} & \textbf{74} & \textbf{76} & \textbf{21} \\ \midrule
2 & inspir.ai robotics                           & 91 & 70 & 71 & 70 
\\ 
3 & VO2021 (Zhao \etal~\cite{zhao2021the}) & 78 & 59 & 69 & 53 
\\ 
4 & Differentiable SLAM-net (\cite{karkus2021differentiable}) & 65 & 47 & 60 & 174
\\
\bottomrule
\end{tabular}
}
\end{center}
\vspace{-5pt}
\caption{\xhdr{Habitat Challenge 2021} benchmark test-standard split (retrieved 2021-Nov-16). The work of `inspir.ai robotics' is concurrent unpublished work.}
\label{table:hc2021}
\vspace{-5pt}
\end{table}

We evaluate our most performant agent (\cref{table:nav_metrics}, row 16) on the Habitat Challenge 2021 benchmark test-std split. Our agent achieves 94\% Success and 74\% SPL (\cref{table:hc2021}) on the test-std split. This is an increase of +16\%~Success/+15\%~SPL over prior published state-of-the-art, Zhao \etal~\cite{zhao2021the}. An unpublished concurrent work increased performance to 91\% Success/70\% SPL and our method improves upon that further.

While our results do not effectively `solve' PointGoal navigation under realistic settings, they improve performance significantly and add more evidence that navigation without building an explicit map \emph{is} possible, even under harsh realistic conditions.

\csection{Real-World Transfer}

Motivated by the strong performance in simulation and by prior work 
establishing strong sim2real predictivity \cite{kadian2019are}, we 
conduct initial zero-shot sim2real experiments  -- \ie, 
deploy our learned agent on a LoCoBot with no sim2real adaptation. 
Across 9 episodes, it achieves 11\%~Success, 65\%~SoftSPL, and makes it 92\% of the way to the goal (SoftSuccess). 
Qualitatively evaluating the the navigation videos (provided on the website \footnote{\href{https://rpartsey.github.io/pointgoalnav}{https://rpartsey.github.io/pointgoalnav}}), the agent does well at avoiding obstacles.
The most challenging part seems to be stopping 
within the success threshold. 
These initial results show promise, and adaptation methods may improve the performance further. 
\csection{Related Work}

Autonomous  navigation  has long  been a subject of research  in  robotics and computer vision~\cite{moravec1984locomotion,durrant-whyte1996localization,nilsson1984shakey}. With advances in computer vision and deep learning, there has been a renewed interest in the use of learning to derive navigation policies for a variety of tasks (such as rearrangement~\cite{szot2021habitat,batra2020rearrangement}, visual navigation,~\cite{anderson2018on,batra2020objectnav,chen2020soundspaces}, and vision-and-language~\cite{anderson2018on,krantz2020beyond}).


\xhdr{Classical vs learned navigators.}
Classical approaches decompose the problem into a sequence of sub-tasks, such as localization, mapping, planning, and control. Each of the sub-tasks is addressed separately and corresponding solutions are then composed into one pipeline. When properly tuned, such methods can perform well.
Wijmans~\etal~\cite{wijmans2020dd} showed that  learned approaches can outperform their classical counterparts with sufficient data and training.

\xhdr{Visual odometry for navigation.}
Given the importance of localization for navigation, 
design choices of the CNN-based relative pose regression given the two consecutive RGB/\rgbd frames and their influence on the downstream navigation metrics of navigation agents has been a subject of prior works \cite{chaplot2020learning,ramakrishnan2020occupancy,datta2020integrating,karkus2021differentiable,zhao2021the}.

Neural SLAM~\cite{chaplot2020learning} integrates learning into classical modular SLAM components and estimates agent pose change by using its predicted egocentric map to update the noisy localization sensor on a LoCoBot.
Built on top of Neural SLAM architecture, Occupancy Anticipation~\cite{ramakrishnan2020occupancy} learns to estimate egomotion directly from \rgbd input and uses egocentric occupancy maps as an auxiliary signal.
Differentiable SLAM-net~\cite{karkus2021differentiable} jointly optimizes all SLAM components by backpropagating through a particle-filter based SLAM algorithm. Such approach significantly improved environment map accuracy that translated into improved downstream navigation performance.

Approaches that do not built an explicit map divide learning agent dynamics and  visual odometry (VO) into two separate components. Initial attempts achieved worse results than approaches that use an explicit map~\cite{datta2020integrating}.
Zhao~\etal~\cite{zhao2021the} focused on improving VO for navigation and showed that map-less approaches can outperform map-building approaches. We continue improvements to VO for navigation and reduce the gap between state-of-the-art performance and an oracle from 31\% SPL to 7\% SPL. 




\csection{Concluding Remarks}

\begin{figure}
    \begin{center}
        \includegraphics[width=0.85\linewidth]{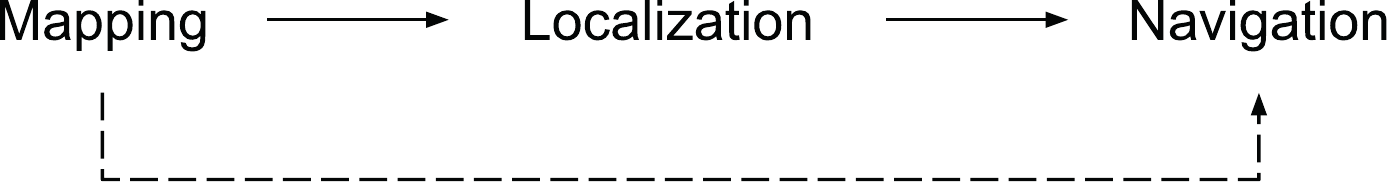}
\end{center}
\vspace{-5pt}
\caption{We study the direct link between mapping and navigation. We find additional evidence that this link is weak. We do not study indirect links however. Indirect links, like the link from mapping to localization to navigation may still be strong.}
\label{fig:dag-mapping-navigation}
\vspace{-5pt}
\end{figure}

We studied the question \myquote{can an autonomous agent navigate in a new environment without building an explicit map?} under (simulated) realistic conditions. Towards answering this question we first demonstrated that when given ground-truth localization (\gpscompass) map-less agents are able to overcome actuation noise and sensor noise  and learn to navigate with near-perfect performance, thereby identifying localization as the limiting factor. 

To improve localization performance, we presented a series of broadly-applicable additions to visual odometry (VO) that improve performance from 64\% Success/52\% SPL to 96\% Success/77\% SPL. While our results do not effectively `solve' PointGoal navigation in the realistic setting, they improve performance significantly and add more evidence that navigation without building an explicit map \emph{is} possible even under harsh realistic conditions.

\xhdr{Limitations.} While our work presents a significant advance in map-less navigation methods for realistic conditions it has several limitations.
\begin{inparaenum}[1)]
\item Embodiment specificity. While our VO model and training procedure are policy agnostic, they are not embodiment agnostic. The importance of action embeddings implies that relaxing this will be challenging, meaning that the VO model may need to be re-trained for each embodiment, which is wasteful.
\item Dataset specificity. Similarly, our learned VO model does not transfer well between datasets and may need to be re-trained for each dataset. We believe large-scale multi-dataset training may be a solution but this remains an open question.
\item Compute requirements. Our best navigation policy used a total of 7.5 billion steps of experience. Training our best VO model required first generating 5M training pairs and then training on 64 GPUs ($\sim$5,000 GPU hours total). High compute requirements were swiftly reduced for PointNav-v1~\cite{ye2020auxiliary,wijmans2020train,shacklett2021large} and we anticipate they will reduce for PointNav-v2 too, but this remains an open direction.
\end{inparaenum}

With regard to the core question, we studied the direct link between mapping and navigation and found increasing evidence that it is a weak link. We have not studied indirect links between mapping and navigation and these may be strong. For instance, there is reason to believe that mapping is needed for accurate localization over long time horizons and localization is needed for navigation (illustrated in \cref{fig:dag-mapping-navigation}). Studying indirect links is an avenue for future work.




\xhdr{Acknowledgements.} The Georgia Tech effort was supported in part by NSF, ONR YIP, and ARO PECASE. EW is supported in part by an ARCS fellowship. The views and conclusions contained herein are those of the authors and should not be interpreted as necessarily representing the official policies or endorsements, either expressed or implied, of the U.S. Government, or any sponsor.

{\small
\bibliographystyle{cvpr2022/ieee_fullname}
\bibliography{bibliography}

\begin{thebibliography}{10}\itemsep=-1pt

\bibitem{anderson2018on}
Peter {Anderson}, Angel~X. {Chang}, Devendra~Singh {Chaplot}, Alexey
  {Dosovitskiy}, Saurabh {Gupta}, Vladlen {Koltun}, Jana {Kosecka}, Jitendra
  {Malik}, Roozbeh {Mottaghi}, Manolis {Savva}, and Amir~Roshan {Zamir}.
\newblock On evaluation of embodied navigation agents.
\newblock {\em arXiv preprint arXiv:1807.06757}, 2018.

\bibitem{ayache1988building}
Nicholas Ayache and Olivier~D Faugeras.
\newblock Building, registrating, and fusing noisy visual maps.
\newblock {\em The International Journal of Robotics Research}, 7(6):45--65,
  1988.

\bibitem{batra2020rearrangement}
Dhruv {Batra}, Angel~X. {Chang}, Sonia {Chernova}, Andrew~J. {Davison}, Jia
  {Deng}, Vladlen {Koltun}, Sergey {Levine}, Jitendra {Malik}, Igor {Mordatch},
  Roozbeh {Mottaghi}, Manolis {Savva}, and Hao {Su}.
\newblock Rearrangement: A challenge for embodied ai.
\newblock {\em arXiv preprint arXiv:2011.01975}, 2020.

\bibitem{batra2020objectnav}
Dhruv {Batra}, Aaron {Gokaslan}, Aniruddha {Kembhavi}, Oleksandr {Maksymets},
  Roozbeh {Mottaghi}, Manolis {Savva}, Alexander {Toshev}, and Erik {Wijmans}.
\newblock Objectnav revisited: On evaluation of embodied agents navigating to
  objects.
\newblock {\em arXiv preprint arXiv:2006.13171}, 2020.

\bibitem{chang2017mp3d}
Angel Chang, Angela Dai, Thomas Funkhouser, Maciej Halber, Matthias Niessner,
  Manolis Savva, Shuran Song, Andy Zeng, and Yinda Zhang.
\newblock Matterport3d: Learning from rgb-d data in indoor environments.
\newblock {\em International Conference on 3D Vision (3DV)}, 2017.

\bibitem{chaplot2020learning}
Devendra~Singh {Chaplot}, Dhiraj {Gandhi}, Saurabh {Gupta}, Abhinav {Gupta},
  and Ruslan {Salakhutdinov}.
\newblock Learning to explore using active neural slam.
\newblock In {\em International Conference on Learning Representations}, 2020.

\bibitem{chen2020soundspaces}
Changan Chen, Unnat Jain, Carl Schissler, Sebastia Vicenc~Amengual Gari, Ziad
  Al-Halah, Vamsi~Krishna Ithapu, Philip Robinson, and Kristen Grauman.
\newblock Soundspaces: Audio-visual navigation in 3d environments.
\newblock In {\em ECCV}, 2020.

\bibitem{choi2015robust}
Sungjoon {Choi}, Qian-Yi {Zhou}, and Vladlen {Koltun}.
\newblock Robust reconstruction of indoor scenes.
\newblock In {\em 2015 IEEE Conference on Computer Vision and Pattern
  Recognition (CVPR)}, pages 5556--5565, 2015.

\bibitem{datta2020integrating}
Samyak {Datta}, Oleksandr {Maksymets}, Judy {Hoffman}, Stefan {Lee}, Dhruv
  {Batra}, and Devi {Parikh}.
\newblock Integrating egocentric localization for more realistic point-goal
  navigation agents.
\newblock {\em arXiv: Computer Vision and Pattern Recognition}, 2020.

\bibitem{durrant-whyte1996localization}
H. Durrant-Whyte, D. Rye, and E. Nebot.
\newblock Localization of autonomous guided vehicles.
\newblock In {\em Robotics Research}, 1996.

\bibitem{gupta2018locobot}
Abhinav Gupta, Adithyavairavan Murali, Dhiraj Gandhi, and Lerrel Pinto.
\newblock Robot learning in homes: Improving generalization and reducing
  dataset bias.
\newblock {\em CoRR}, abs/1807.07049, 2018.

\bibitem{he2016deep}
Kaiming {He}, Xiangyu {Zhang}, Shaoqing {Ren}, and Jian {Sun}.
\newblock Deep residual learning for image recognition.
\newblock In {\em 2016 IEEE Conference on Computer Vision and Pattern
  Recognition (CVPR)}, pages 770--778, 2016.

\bibitem{hochreiter1997long}
Sepp {Hochreiter} and Jürgen {Schmidhuber}.
\newblock Long short-term memory.
\newblock {\em Neural Computation}, 9(8):1735--1780, 1997.

\bibitem{ioffe2015batch}
Sergey {Ioffe} and Christian {Szegedy}.
\newblock Batch normalization: Accelerating deep network training by reducing
  internal covariate shift.
\newblock In {\em Proceedings of The 32nd International Conference on Machine
  Learning}, volume~1, pages 448--456, 2015.

\bibitem{kadian2019are}
Abhishek {Kadian}, Joanne {Truong}, Aaron {Gokaslan}, Alexander {Clegg}, Erik
  {Wijmans}, Stefan {Lee}, Manolis {Savva}, Sonia {Chernova}, and Dhruv
  {Batra}.
\newblock Are we making real progress in simulated environments? measuring the
  sim2real gap in embodied visual navigation.
\newblock {\em arXiv: Computer Vision and Pattern Recognition}, 2019.

\bibitem{karkus2021differentiable}
Péter {Karkus}, Shaojun {Cai}, and David {Hsu}.
\newblock Differentiable slam-net: Learning particle slam for visual
  navigation.
\newblock In {\em Proceedings of the IEEE/CVF Conference on Computer Vision and
  Pattern Recognition}, pages 2815--2825, 2021.

\bibitem{krantz2020beyond}
Jacob {Krantz}, Erik {Wijmans}, Arjun {Majumdar}, Dhruv {Batra}, and Stefan
  {Lee}.
\newblock Beyond the nav-graph: Vision-and-language navigation in continuous
  environments.
\newblock In {\em European Conference on Computer Vision}, pages 104--120,
  2020.

\bibitem{krizhevsky2017imagenet}
Alex {Krizhevsky}, Ilya {Sutskever}, and Geoffrey~E. {Hinton}.
\newblock Imagenet classification with deep convolutional neural networks.
\newblock {\em Communications of The ACM}, 60(6):84--90, 2017.

\bibitem{moravec1984locomotion}
Hans Moravec.
\newblock Locomotion, vision and intelligence.
\newblock In Michael Brady and Richard Paul, editors, {\em Proceedings of
  Robotics Research - The First International Symposium}, pages 215--224. MIT
  Press, August 1984.

\bibitem{murali2019pyrobot}
Adithyavairavan {Murali}, Tao {Chen}, Kalyan~Vasudev {Alwala}, Dhiraj {Gandhi},
  Lerrel {Pinto}, Saurabh {Gupta}, and Abhinav {Gupta}.
\newblock Pyrobot: An open-source robotics framework for research and
  benchmarking.
\newblock {\em arXiv: Robotics}, 2019.

\bibitem{nilsson1984shakey}
N Nilsson.
\newblock Shakey the robot, 1984.

\bibitem{okeefe1978hippocampus}
John O'keefe and Lynn Nadel.
\newblock {\em The hippocampus as a cognitive map}.
\newblock Oxford: Clarendon Press, 1978.

\bibitem{NEURIPS2019_9015}
Adam Paszke, Sam Gross, Francisco Massa, Adam Lerer, James Bradbury, Gregory
  Chanan, Trevor Killeen, Zeming Lin, Natalia Gimelshein, Luca Antiga, Alban
  Desmaison, Andreas Kopf, Edward Yang, Zachary DeVito, Martin Raison, Alykhan
  Tejani, Sasank Chilamkurthy, Benoit Steiner, Lu Fang, Junjie Bai, and Soumith
  Chintala.
\newblock Pytorch: An imperative style, high-performance deep learning library.
\newblock In H. Wallach, H. Larochelle, A. Beygelzimer, F. d\textquotesingle
  Alch\'{e}-Buc, E. Fox, and R. Garnett, editors, {\em Advances in Neural
  Information Processing Systems 32}, pages 8024--8035. Curran Associates,
  Inc., 2019.

\bibitem{ramakrishnan2020occupancy}
Santhosh~K. {Ramakrishnan}, Ziad {Al-Halah}, and Kristen {Grauman}.
\newblock Occupancy anticipation for efficient exploration and navigation.
\newblock In {\em ECCV (5)}, pages 400--418, 2020.

\bibitem{ramakrishnan2021habitat}
Santhosh~K. {Ramakrishnan}, Aaron {Gokaslan}, Erik {Wijmans}, Oleksandr
  {Maksymets}, Alexander {Clegg}, John {Turner}, Eric {Undersander}, Wojciech
  {Galuba}, Andrew {Westbury}, Angel~X. {Chang}, Manolis {Savva}, Yili {Zhao},
  and Dhruv {Batra}.
\newblock Habitat-matterport 3d dataset (hm3d): 1000 large-scale 3d
  environments for embodied ai.
\newblock {\em arXiv preprint arXiv:2109.08238}, 2021.

\bibitem{savva2019habitat}
Manolis {Savva}, Jitendra {Malik}, Devi {Parikh}, Dhruv {Batra}, Abhishek
  {Kadian}, Oleksandr {Maksymets}, Yili {Zhao}, Erik {Wijmans}, Bhavana {Jain},
  Julian {Straub}, Jia {Liu}, and Vladlen {Koltun}.
\newblock Habitat: A platform for embodied ai research.
\newblock In {\em 2019 IEEE/CVF International Conference on Computer Vision
  (ICCV)}, pages 9339--9347, 2019.

\bibitem{shacklett2021large}
Brennan Shacklett, Erik Wijmans, Aleksei Petrenko, Manolis Savva, Dhruv Batra,
  Vladlen Koltun, and Kayvon Fatahalian.
\newblock Large batch simulation for deep reinforcement learning.
\newblock {\em Int. Conf. Learn. Represent.}, 2021.

\bibitem{smith1990estimating}
Randall Smith, Matthew Self, and Peter Cheeseman.
\newblock Estimating uncertain spatial relationships in robotics.
\newblock In {\em Autonomous robot vehicles}, pages 167--193. Springer, 1990.

\bibitem{srivastava2014dropout}
Nitish {Srivastava}, Geoffrey {Hinton}, Alex {Krizhevsky}, Ilya {Sutskever},
  and Ruslan {Salakhutdinov}.
\newblock Dropout: a simple way to prevent neural networks from overfitting.
\newblock {\em Journal of Machine Learning Research}, 15(1):1929--1958, 2014.

\bibitem{szot2021habitat}
Andrew {Szot}, Alexander {Clegg}, Eric {Undersander}, Erik {Wijmans}, Yili
  {Zhao}, John {Turner}, Noah {Maestre}, Mustafa {Mukadam}, Devendra~Singh
  {Chaplot}, Oleksandr {Maksymets}, Aaron {Gokaslan}, Vladimir {Vondrus},
  Sameer {Dharur}, Franziska {Meier}, Wojciech {Galuba}, Angel {Chang}, Zsolt
  {Kira}, Vladlen {Koltun}, Jitendra {Malik}, Manolis {Savva}, and Dhruv
  {Batra}.
\newblock Habitat 2.0: Training home assistants to rearrange their habitat.
\newblock {\em arXiv preprint arXiv:2106.14405}, 2021.

\bibitem{thrun2005probabilistic}
Sebastian Thrun, Wolfram Burgard, and Dieter Fox.
\newblock Probabilistic robotics (intelligent robotics and autonomous agents),
  2005.

\bibitem{tolman48}
Edward~C. Tolman.
\newblock Cognitive maps in rats and men.
\newblock {\em Psychological Review}, 55(4):189--208, 1948.

\bibitem{wijmans2020train}
Erik Wijmans, Irfan Essa, and Dhruv Batra.
\newblock How to train pointgoal navigation agents on a (sample and compute)
  budget.
\newblock {\em arXiv preprint arXiv:2012.06117}, 2020.

\bibitem{wijmans2020dd}
Erik {Wijmans}, Abhishek {Kadian}, Ari {Morcos}, Stefan {Lee}, Irfan {Essa},
  Devi {Parikh}, Manolis {Savva}, and Dhruv {Batra}.
\newblock Dd-ppo: Learning near-perfect pointgoal navigators from 2.5 billion
  frames.
\newblock In {\em Eighth International Conference on Learning Representations},
  2020.

\bibitem{wu2018group}
Yuxin {Wu} and Kaiming {He}.
\newblock Group normalization.
\newblock {\em arXiv: Computer Vision and Pattern Recognition}, 2018.

\bibitem{xia2018gibson}
Fei {Xia}, Amir~R. {Zamir}, Zhiyang {He}, Alexander {Sax}, Jitendra {Malik},
  and Silvio {Savarese}.
\newblock Gibson env: Real-world perception for embodied agents.
\newblock In {\em 2018 IEEE/CVF Conference on Computer Vision and Pattern
  Recognition}, pages 9068--9079, 2018.

\bibitem{ye2020auxiliary}
Joel Ye, Dhruv Batra, Erik Wijmans, and Abhishek Das.
\newblock Auxiliary tasks speed up learning pointgoal navigation.
\newblock {\em Conference on Robot Learning (CoRL)}, 2020.

\bibitem{sct21iros}
Naoki Yokoyama, Sehoon Ha, and Dhruv Batra.
\newblock Success weighted by completion time: {A} dynamics-aware evaluation
  criteria for embodied navigation.
\newblock In {\em 2021 IEEE/RSJ International Conference on Intelligent Robots
  and Systems (IROS)}, 2021.

\bibitem{yun2019cutmix}
Sangdoo {Yun}, Dongyoon {Han}, Sanghyuk {Chun}, Seong~Joon {Oh}, Youngjoon
  {Yoo}, and Junsuk {Choe}.
\newblock Cutmix: Regularization strategy to train strong classifiers with
  localizable features.
\newblock In {\em 2019 IEEE/CVF International Conference on Computer Vision
  (ICCV)}, pages 6022--6031, 2019.

\bibitem{zhang2017mixup}
Hongyi {Zhang}, Moustapha {Cisse}, Yann~N. {Dauphin}, and David {Lopez-Paz}.
\newblock mixup: Beyond empirical risk minimization.
\newblock In {\em International Conference on Learning Representations}, 2017.

\bibitem{zhao2021the}
Xiaoming {Zhao}, Harsh {Agrawal}, Dhruv {Batra}, and Alexander~G. {Schwing}.
\newblock The surprising effectiveness of visual odometry techniques for
  embodied pointgoal navigation.
\newblock In {\em Proceedings of the IEEE/CVF International Conference on
  Computer Vision}, pages 16127--16136, 2021.

\end{thebibliography}
}

\clearpage


\onecolumn
\section{Supplementary}

\subsection{VO Prediction Error}
We report Mean Absolute Error (MAE) \cref{eq:mse_metric} between ground-truth and estimated egomotion  in \cref{table:visual_odometry_metrics}.

\begin{equation}
\label{eq:mse_metric}
MAE = \frac{1}{N}\sum_{i=1}^{N}\left( |x - \hat{x}| +|y - \hat{y}| +|z - \hat{z}| \right) \\ + \frac{1}{N}\sum_{i=1}^{N}|\theta - \hat{\theta}| 
\end{equation}

\begin{table*}[ht!]
\begin{center}
\resizebox{\textwidth}{!}{
\begin{tabular}{ccccccccccccccccc}
\toprule
\multicolumn{1}{c}{} & 
\multicolumn{1}{c}{Dataset} &
\multicolumn{2}{c}{VO} &
\multicolumn{2}{c}{Embedding} &
\multicolumn{2}{c}{Train time} & 
\multicolumn{1}{c}{\multirow{2}{*}{Epoch}} &
\multicolumn{4}{c}{Translation MAE (cm)} & 
\multicolumn{4}{c}{Rotation MAE (centi-radians)} 
\\ 
\cmidrule(lr){3-4} \cmidrule(lr){5-6} \cmidrule(lr){7-8} \cmidrule(lr){10-13} \cmidrule(l){14-17}
\multicolumn{1}{l}{} &
\multicolumn{1}{c}{size(M)} & 
\multicolumn{1}{c}{Encoder} & 
\multicolumn{1}{c}{Size(M)} & 
\multicolumn{1}{c}{1FC} &
\multicolumn{1}{c}{2FC} &
\multicolumn{1}{c}{Flip} &
\multicolumn{1}{c}{Swap} &
\multicolumn{1}{c}{} & 
\multicolumn{1}{c}{Total} & 
\multicolumn{1}{c}{Forward} & 
\multicolumn{1}{c}{Left} & 
\multicolumn{1}{c}{Right} & 
\multicolumn{1}{c}{Total} & 
\multicolumn{1}{c}{Forward} & 
\multicolumn{1}{c}{Left} & 
\multicolumn{1}{c}{Right}   
\\
\midrule
1 & 0.5 & ResNet18 & 4.2 &  & & &  & 50 & 2.65 & 2.21 & 3.14 & 3.30 & 1.00 & 0.66 & 1.41 & 1.45 \\ \midrule
2 & 0.5 & ResNet18 & 4.2 & \checkmark & & &  & 43 & 2.45 & 1.82 & 3.17 & 3.37 & 0.90 & 0.56 & 1.30 & 1.39 \\ \midrule
3 & 0.5 & ResNet18 & 4.2 & \checkmark & \checkmark & & & 44 & 2.38 & 1.78 & 3.08 & 3.22 & 0.86 & 0.55 & 1.23 & 1.31 \\ \midrule
4 & 0.5 & ResNet18 & 4.2 & \checkmark & \checkmark & & \checkmark & 48 & 2.60 & 2.22 & 3.06 & 3.12 & 0.90 & 0.66 & 1.21 & 1.2 \\ \midrule
5 & 0.5 & ResNet18 & 4.2 & \checkmark & \checkmark & \checkmark & & 50 & 2.26 & 1.77 & 2.86 & 2.92 & 0.75 & 0.49 & 1.09 & 1.1 \\ \midrule
6 & 0.5 & ResNet18 & 4.2 & \checkmark & \checkmark & \checkmark & \checkmark & 50 & 2.26 & 2.02 & 2.56 & 2.56 & 0.75 & 0.55 & 0.98 & 1.03 \\ \midrule
7 & 1.5 & ResNet18 & 4.2 & \checkmark & \checkmark & & & 48 & 1.94 & 1.33 & 2.72 & 2.71 & 0.69 & 0.43 & 1.03 & 1.02 \\  \midrule
8 & 1.5 & ResNet18 & 4.2 & \checkmark & \checkmark & \checkmark & \checkmark & 50 & 1.7 & 1.48 & 1.94 & 2.02 & 0.61 & 0.48 & 0.75 & 0.81 \\ \midrule
9 & 1.5 & ResNet50 & 7.6 & \checkmark & \checkmark & \checkmark & \checkmark & 48 & 1.48 & 1.34 & 1.63 & 1.68 & 0.49 & 0.38 & 0.57 & 0.67 \\ 
\bottomrule
\end{tabular}
}
\caption{Mean Absolute Error (MAE) between ground-truth and estimated egomotion for all types of actions in total and for each type of actions separately (corresponding to \cref{table:nav_metrics}).} \label{table:visual_odometry_metrics}
\end{center}
\end{table*}

\subsection{Qualitative Results}
\begin{figure}[ht]
    \begin{center}
        \includegraphics[width=0.95\linewidth]{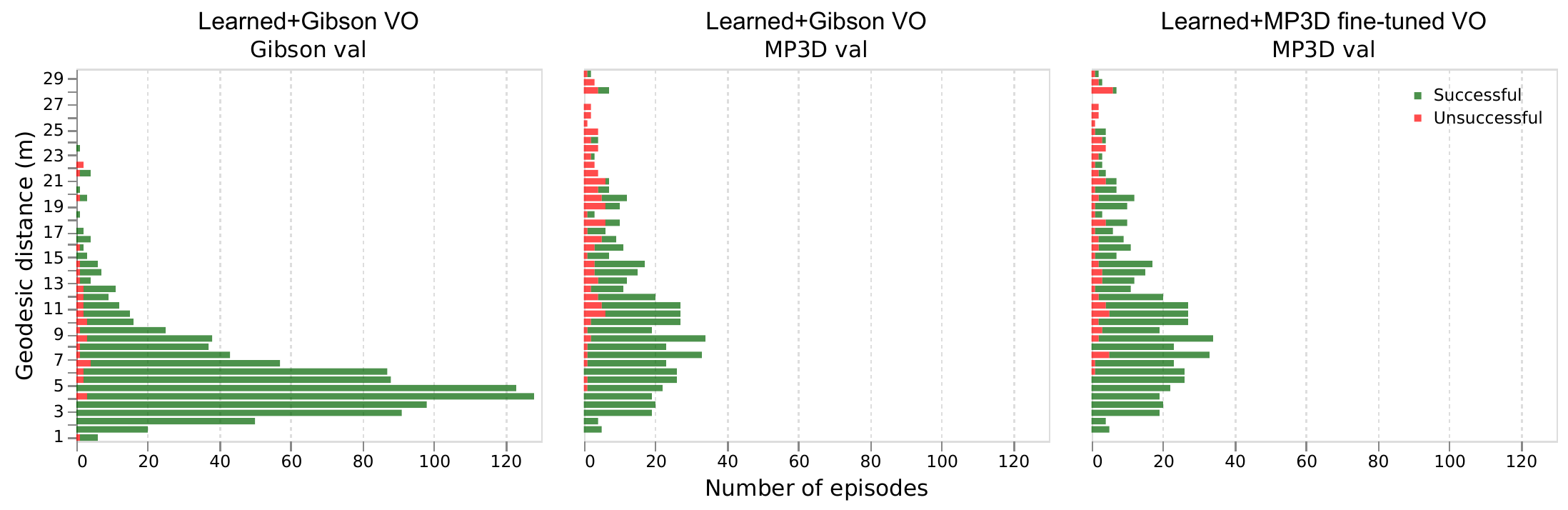}
\end{center}
\caption{Success vs.~path length. Our method performs worse on longer episodes. Fine-tuning on MP3D improves performance.}
\label{fig:path-length-comp}
\end{figure}

\begin{figure*}[!ht]
    \begin{center}
        \includegraphics[width=1\linewidth]{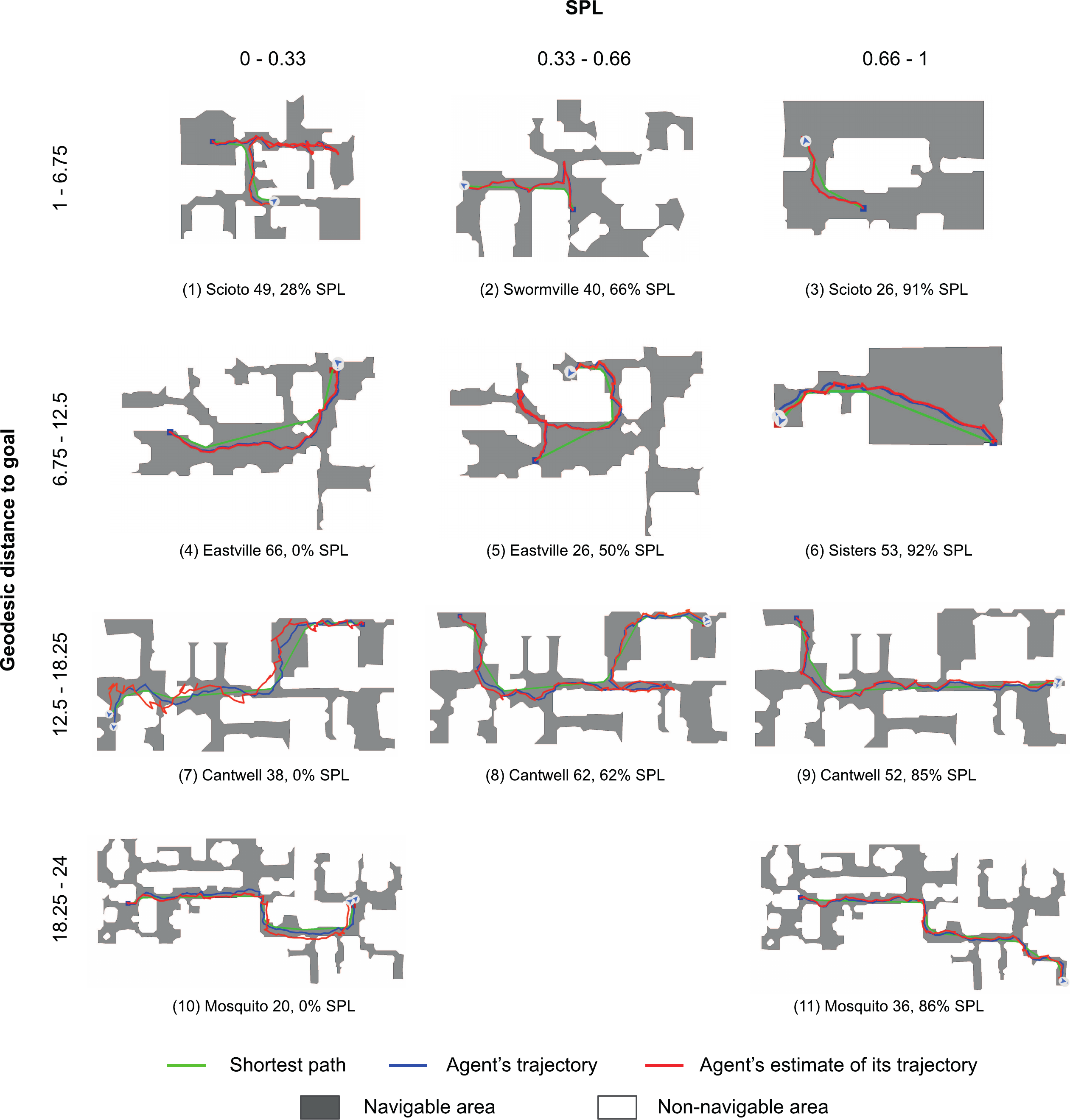}
\end{center}
\caption{Our best HC 2021 PointNav agent's (row 16 in \cref{table:nav_metrics}) navigation trajectories on Gibson 4+ val scenes (and Gibson-v2 PointGoal navigation episodes) broken down by geodesic distance between agent’s spawn location and target (on rows) vs SPL achieved by the agent (on cols). The color of the trajectory changes from dark to light over time (\texttt{cv2.COLORMAP\_WINTER} for agent's trajectory, \texttt{cv2.COLORMAP\_AUTUMN} for agent's estimate of its trajectory).
}
\label{fig:top-down-maps-gibson}
\end{figure*}

\begin{figure*}[!ht]
    \begin{center}
        \includegraphics[width=1\linewidth]{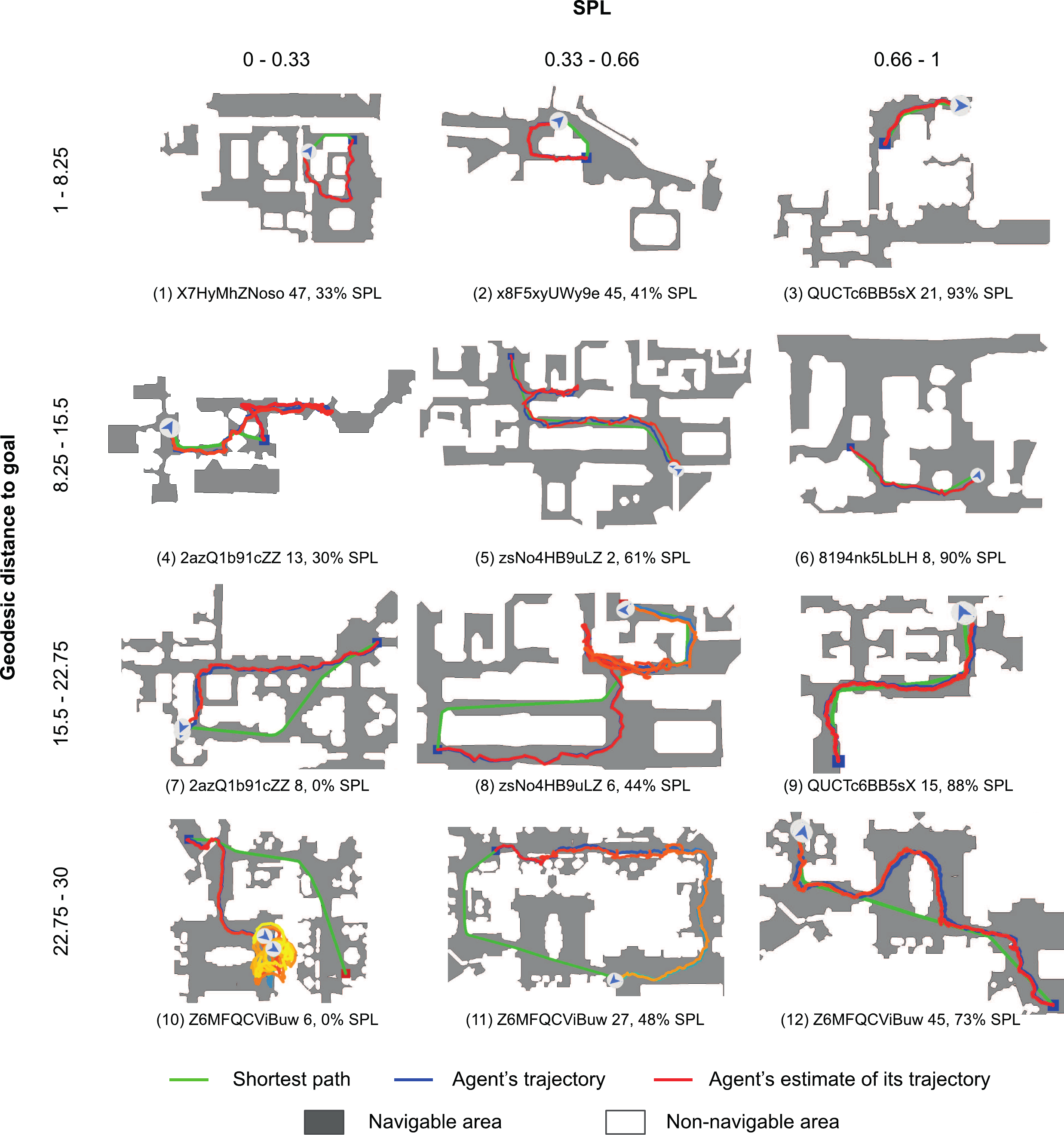}
\end{center}
\caption{Our best HC 2021 PointNav agent's (row 16 in \cref{table:nav_metrics}) navigation trajectories on MP3D val scenes (and MP3D-v2 PointGoal navigation episodes) broken down by geodesic distance between agent’s spawn location and target (on rows) vs SPL achieved by the agent (on cols). The color of the trajectory changes from dark to light over time (\texttt{cv2.COLORMAP\_WINTER} for agent's trajectory, \texttt{cv2.COLORMAP\_AUTUMN} for agent's estimate of its trajectory).
}
\label{fig:top-down-maps-mp3d}
\end{figure*}

Complementary to \cref{table:cross-dataset-eval} we provide additional qualitative results when integrating the navigation policy with our VO model. On Gibson 4+ val scenes dataset (navigation trajectories illustrated in \cref{fig:top-down-maps-gibson}) our navigation agent follows a near-perfect path for both: episodes with a relatively small geodesic distance to the target (row~1 -- row~2) and episodes with a large geodesic distance to the target (row~3 -- row~4) and can do backtracking when the wrong way was chosen (top-down maps (1), (2), (5) and (8)). We found that on Matterport3D (MP3D) (navigation trajectories illustrated in \cref{fig:top-down-maps-mp3d}) the navigation performance suffers (see \cref{table:cross-dataset-eval}). First of all, the Matterport3D navigation episodes are `longer': $10.92m$ average geodesic distance to goal on MP3D vs $5.89m$ average geodesic distance to goal on Gibson 4+. Larger scenes usually have more than one way to the target place and if agent chooses longer way it affects navigation metrics (top-down maps (1), (2), (5) and (8)). We also noticed that it is harder for agent to backtrack in larger scenes (top-down map (10)).

To show that our method isn't specific to Gibson, we have fine-tuned our VO model on MP3D and increased performance to
85.6~Success/65.9~SPL. While this is still worse than our Gibson performance, note that MP3D has harder and longer navigation episodes (\cref{fig:path-length-comp}) and this result is  better than prior SOTA on Gibson (71\% Success).

\clearpage
\section{Simulation-to-reality Transfer}
\label{section:sim2real}

\begin{table}[ht]
\begin{center}
\resizebox{\linewidth}{!}{
\begin{tabular}{@{}lcccccccccc@{}}
\toprule
\multicolumn{1}{c}{\multirow{2}{*}{Episode name}} &
\multicolumn{1}{c}{\multirow{2}{*}{Run}} &
\multicolumn{2}{c}{Path length (m)} &
\multicolumn{3}{c}{Navigation metrics} &
\multicolumn{4}{c}{Success} 
\\
\cmidrule(l){3-4}
\cmidrule(l){5-7}
\cmidrule(l){8-11}
\multicolumn{2}{c}{} & 
\multicolumn{1}{c}{Shortest} & 
\multicolumn{1}{c}{Agent's} & 
\multicolumn{1}{c}{$\text{d}_\text{G}$} &
\multicolumn{1}{c}{SoftSuccess} &
\multicolumn{1}{c}{SoftSPL} &
\multicolumn{1}{c}{$\text{d}_\text{G} < 0.36$} &
\multicolumn{1}{c}{$\text{d}_\text{G} < 0.395$} &
\multicolumn{1}{c}{$\text{d}_\text{G} < 0.45$} &
\multicolumn{1}{c}{$\text{d}_\text{G} < 0.70$} 
\\
\midrule
kitchen2couch & 1 & 4.94 & 7.04 & 0.39 & 0.92 & 0.70 & 0 & 1 & 1 & 1 \\
kitchen2couch & 2 & 4.94 & 9.06 & 0.44 & 0.91 & 0.54 & 0 & 0 & 1 & 1 \\
kitchen2couch & 3 & 4.94 & 7.69 & 0.25 & 0.95 & 0.64 & 1 & 1 & 1 & 1  
\\ 
\midrule
desk2bathroom & 1 & 8.37 & 11.26 & 0.38 & 0.96 & 0.74 & 0 & 1 & 1 & 1 \\
desk2bathroom & 2 & 8.37 & 12.32 & 0.69 & 0.92 & 0.68 & 0 & 0 & 0 & 1 \\ 				
desk2bathroom & 3 & 8.37 &  9.01 & 0.76 & 0.91 & 0.93 & 0 & 0 & 0 & 0 				
\\ 
\midrule
bed2desk & 1 & 8.23 & 11.82 & 0.65 & 0.92 & 0.70 & 0 & 0 & 0 & 1 \\
bed2desk & 2 & 8.23 & 10.77 & 1.00 & 0.88 & 0.76 & 0 & 0 & 0 & 0 \\ 				
bed2desk & 3 & 8.23 & 11.98 & 0.60 & 0.93 & 0.69 & 0 & 0 & 0 & 1 	
\\ 
\midrule
 & & & Average: & 0.57 & 0.92 & 0.71 & 0.11 & 0.33 & 0.44 & 0.78
\\
\bottomrule
\end{tabular}
}
\end{center}
\vspace{-5pt}
\caption{\xhdr{Sim2real transfer.} The shortest path length was calculated using RRT*~\cite{sct21iros} for 5000 iterations.} \label{table:sim2real-eval}
\vspace{-5pt}
\end{table}

\xhdr{Policy Training.} We use a similar training recipe as in the main paper to train our policy with the follow modifications. We alter the simulated camera config (e.g., FOV, mounting height) to match the Intel RealSense D435 Depth camera attached on our robot. Additionally, we use all of Gibson~\cite{xia2018gibson}, MP3D~\cite{chang2017mp3d}, and HM3D~\cite{ramakrishnan2021habitat} for training scenes to increase diversity.

\xhdr{Visual Odometery Training.} For visual odometry training, we again modify the camera config to match both the RGB and Depth cameras of the Intel RealSense D435. Note that we use the unaligned RGB and Depth output from the camera (i.e., we do not do a re-centered crop on the Depth images) to leverage the fact that the Depth camera's field-of-view is larger than that of the RGB camera and increase the amount of overlap between observations after a rotation action.

\xhdr{Deployment details.} We use a LoCoBot~\cite{gupta2018locobot} with the Kobuki base. We mount an Intel RealSense D435 camera 0.61 meters from the ground with a tilt angle of 0 degrees. The policy only uses the RGB and Depth images from the RealSense camera as input. The Depth images were de-noised using a median filter. Mapping and localization using the robot's LiDAR are used solely for visualization and calculating the lengths of the shortest path and the agent's path. 

\begin{figure*}[!ht]
    \begin{center}
        \includegraphics[width=1\linewidth]{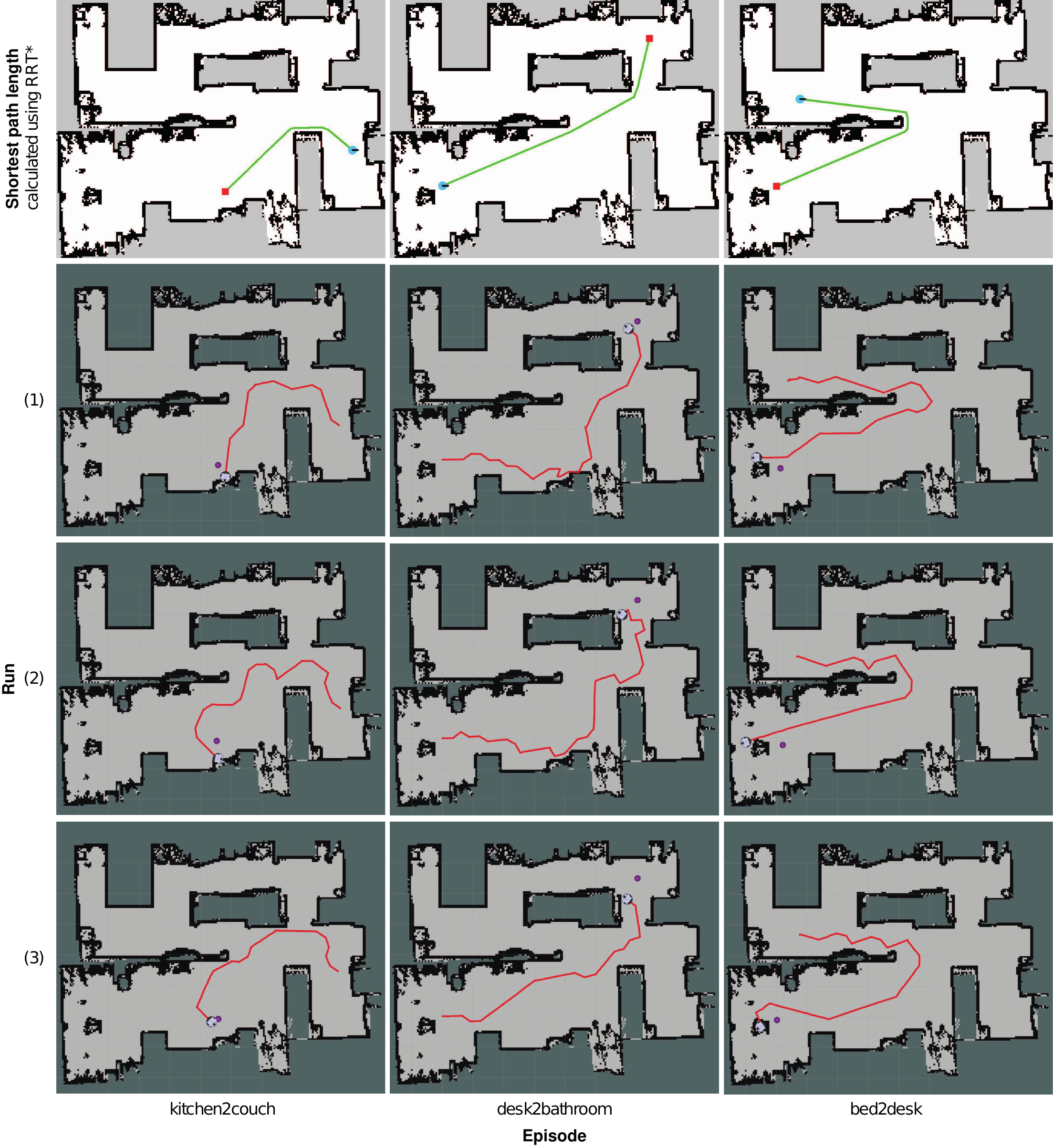}
\end{center}
\caption{Reality experiments top-down maps.
}
\label{fig:top-down-maps-reality}
\end{figure*}

We present the results of zero-shot sim2real experiments in \cref{table:sim2real-eval} and corresponding robot's navigation trajectories  in \cref{fig:top-down-maps-reality}.

\end{document}